\def\year{2020}\relax
\documentclass[letterpaper]{article}

\usepackage{aaai20}
\usepackage{times}
\usepackage{helvet}
\usepackage{courier}
\usepackage{mathtools}

\usepackage[hyphens]{url}  
\usepackage[utf8]{inputenc} 
\usepackage{booktabs}       
\usepackage{amsfonts}       
\usepackage{nicefrac}       
\usepackage{microtype}      
\usepackage{paralist}
\usepackage{microtype}
\usepackage{graphicx}
\usepackage{subfigure}
\usepackage{booktabs} 
\usepackage{amsfonts,amsmath,amssymb,amsthm}
\usepackage{bm,nicefrac}
\usepackage{graphicx}
\usepackage{mathtools}
\usepackage{lineno}
\usepackage{lipsum}

\usepackage{array}
\usepackage{multirow}

\usepackage{floatrow}
\newfloatcommand{capbtabbox}{table}[][\FBwidth]
\newcommand{\STAB}[1]{\begin{tabular}{@{}c@{}}#1\end{tabular}}

\usepackage[]{graphicx,amssymb}    
\usepackage{amsmath,bm}
\usepackage{epsfig}
\usepackage{subfigure}
\usepackage{epstopdf}
\usepackage{color} 

\DeclareMathOperator*{\E}{\mathbb{E}}

\usepackage[commentmarkup=footnote,highlightmarkup=uwave]{changes}
\definechangesauthor[name={FC}, color=orange]{fc}
\definechangesauthor[name={MS}, color=red]{ms}
\definechangesauthor[name={LK}, color=blue]{lk}

\title{Uncertainty-Aware Deep Classifiers using Generative Models}

\author{Murat Sensoy,\textsuperscript{\rm 1,2}
Lance Kaplan,\textsuperscript{\rm 3}
Federico Cerutti,\textsuperscript{\rm 4,5}
Maryam Saleki\textsuperscript{\rm 2}\\
\textsuperscript{1}{Blue Prism AI Labs, London, UK}\\
\textsuperscript{2}{Department of Computer Science, Ozyegin University, Istanbul, Turkey}\\
\textsuperscript{3}{US Army Research Lab, Adelphi, MD 20783, USA}\\
\textsuperscript{4}{Department of Information Engineering, University of Brescia, 25123 Brescia, Italy}\\
\textsuperscript{5}{Cardiff University, Cardiff, CF10 3AT, UK}\\
murat.sensoy@blueprism.com,
lkaplan@ieee.org, 
federico.cerutti@unibs.it, maryam.saleki@ozu.edu.tr}

\begin{document}

\maketitle

\begin{abstract}
Deep neural networks are often ignorant about what they do not know and overconfident when they make uninformed predictions.
Some recent approaches quantify classification uncertainty directly by training the model to output high uncertainty for the data samples close to class boundaries or from the outside of the training distribution.
These approaches use an auxiliary data set during training to represent out-of-distribution samples.
However, selection or creation of such an auxiliary data set is non-trivial, especially for high dimensional data such as images. In this work we develop a novel neural network model that is able to express both aleatoric and epistemic uncertainty to distinguish decision boundary and out-of-distribution regions of the feature space. To this end, variational autoencoders and generative adversarial networks are incorporated to automatically generate out-of-distribution exemplars for training.
%
%
%
Through extensive analysis, we demonstrate that the proposed approach provides better estimates of uncertainty for in- and out-of-distribution samples, and adversarial examples on well-known data sets against state-of-the-art approaches including recent Bayesian approaches for neural networks and anomaly detection methods.
\end{abstract}

\section{Introduction}
\label{sec:intro}
%
While deep learning models demonstrate remarkable generalization performance in light of the large number of parameters they exploit, they can be misleadingly overconfident when they do make mistakes. 
%
%
%
The false sense of trust these models create may have serious consequences, especially if they are used for high-risk tasks.
A striking example is the misclassification of the white side of a trailer as bright sky: this caused a car operating with automated vehicle control
systems to crash against a tractor-semitrailer truck near Williston,
Florida, USA on 7th May 2016. The car driver died due to the sustained injury~\cite{NHTSA}.
%

There are two categories of uncertainty \cite{Matthies2007}.
%
\emph{Epistemic uncertainty}, or \emph{model uncertainty}, results from  limited knowledge and could in principle be reduced: uncertain predictions for out-of-distribution samples fall into this category.
Among other approaches, Bayesian deep learning methods try to estimate epistemic uncertainty by modeling the distributions for the parameters values, distributions that seldom admit closed-form representations, hence requiring expensive Monte Carlo sampling methods~\cite{bishop2006pattern}.

\emph{Aleatoric uncertainty}, or \emph{data uncertainty}, is the noise inherent in the observations (e.g., label noise), or class overlap: unlike epistemic uncertainty, aleatoric uncertainty cannot be reduced by observing more data samples. For instance, having identical samples with different labels, e.g., on the class boundary, is an example of aleatoric uncertainty.
Approaches such as Evidential Neural Networks (EDL) and Lightweight Probabilistic Deep Networks, are proposed recently to estimate aleatoric uncertainties in deep neural networks by directly estimating parameters of the predictive posterior as their output~\cite{edl,gast2018lightweight}.
These approaches do not require any sampling and assume minimal changes to the architecture of standard neural networks.

\begin{figure}
  \begin{tabular}{cc}
      \includegraphics[width=3.0cm]{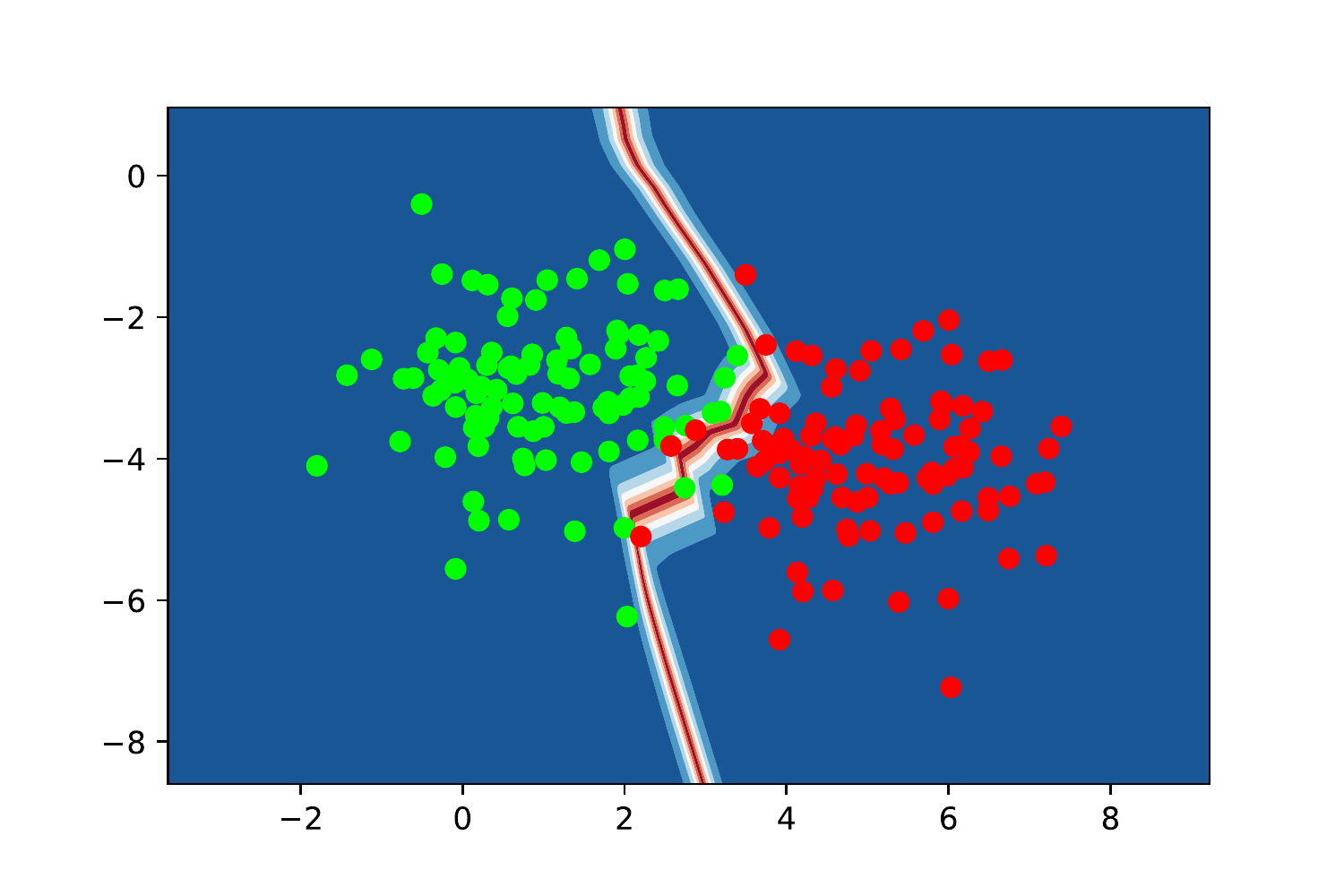} &
      \includegraphics[width=3.0cm]{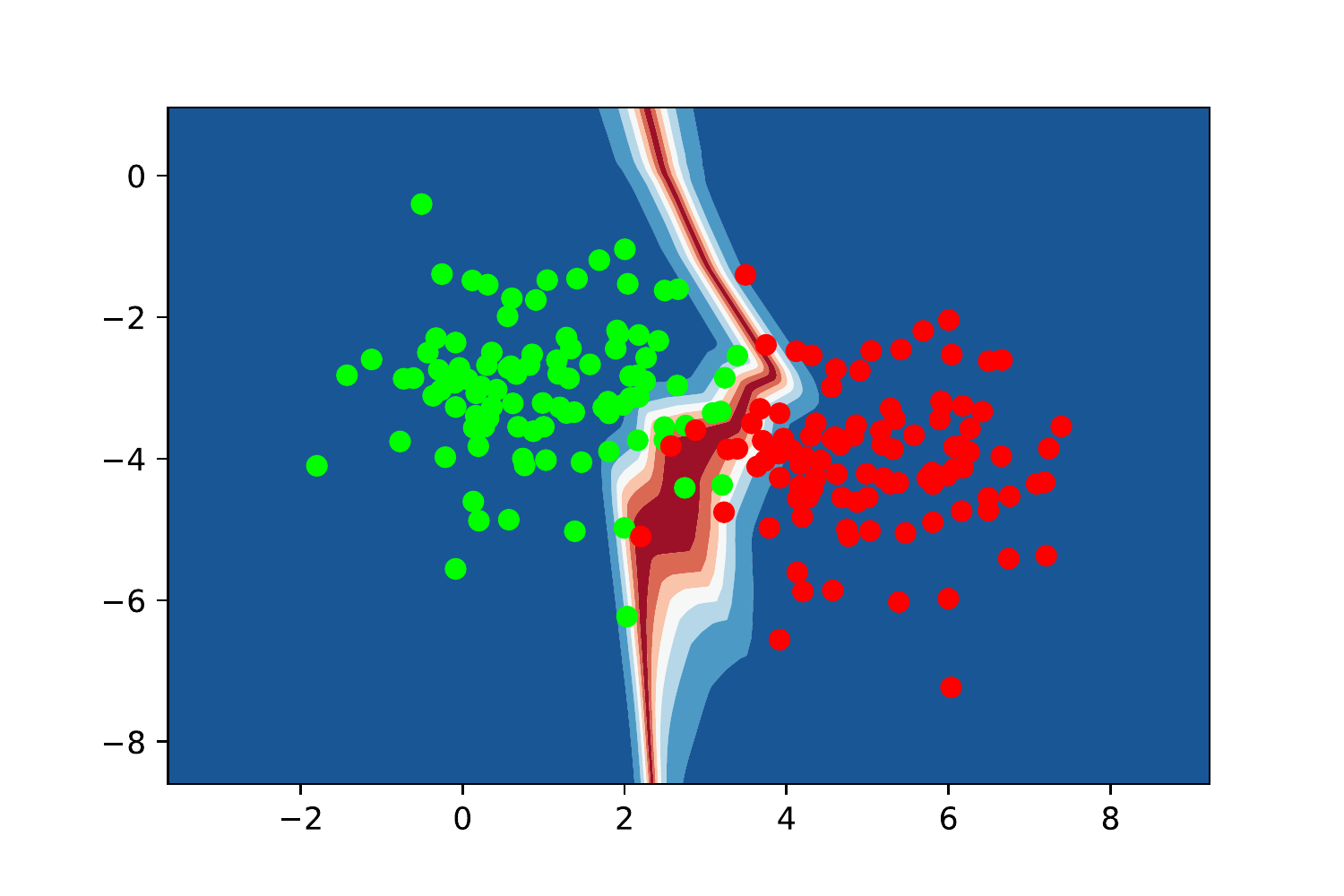} \\
      (a) Standard Nets & (b) Evidential Nets \\
       \includegraphics[width=3.0cm]{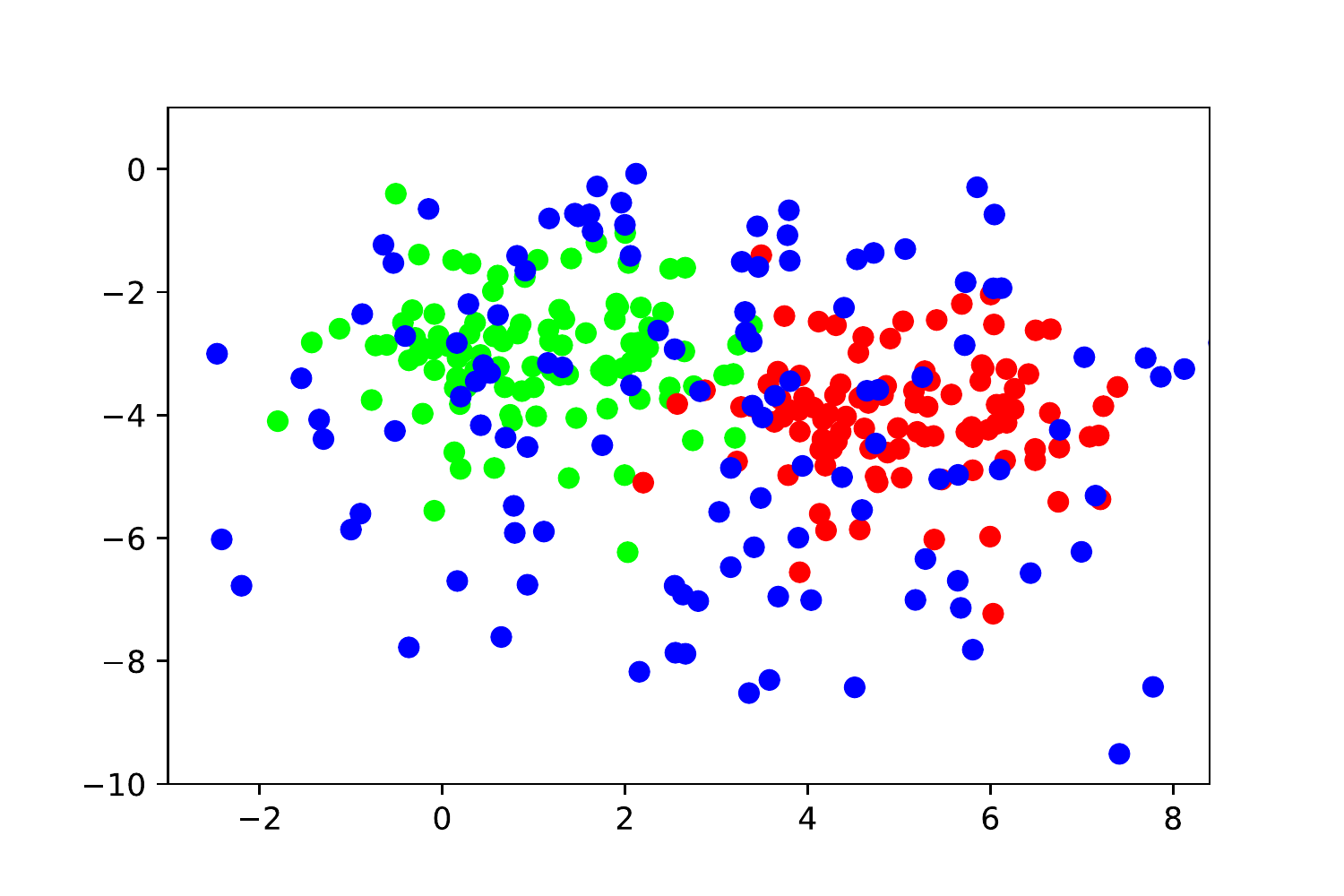} & \includegraphics[width=3.0cm]{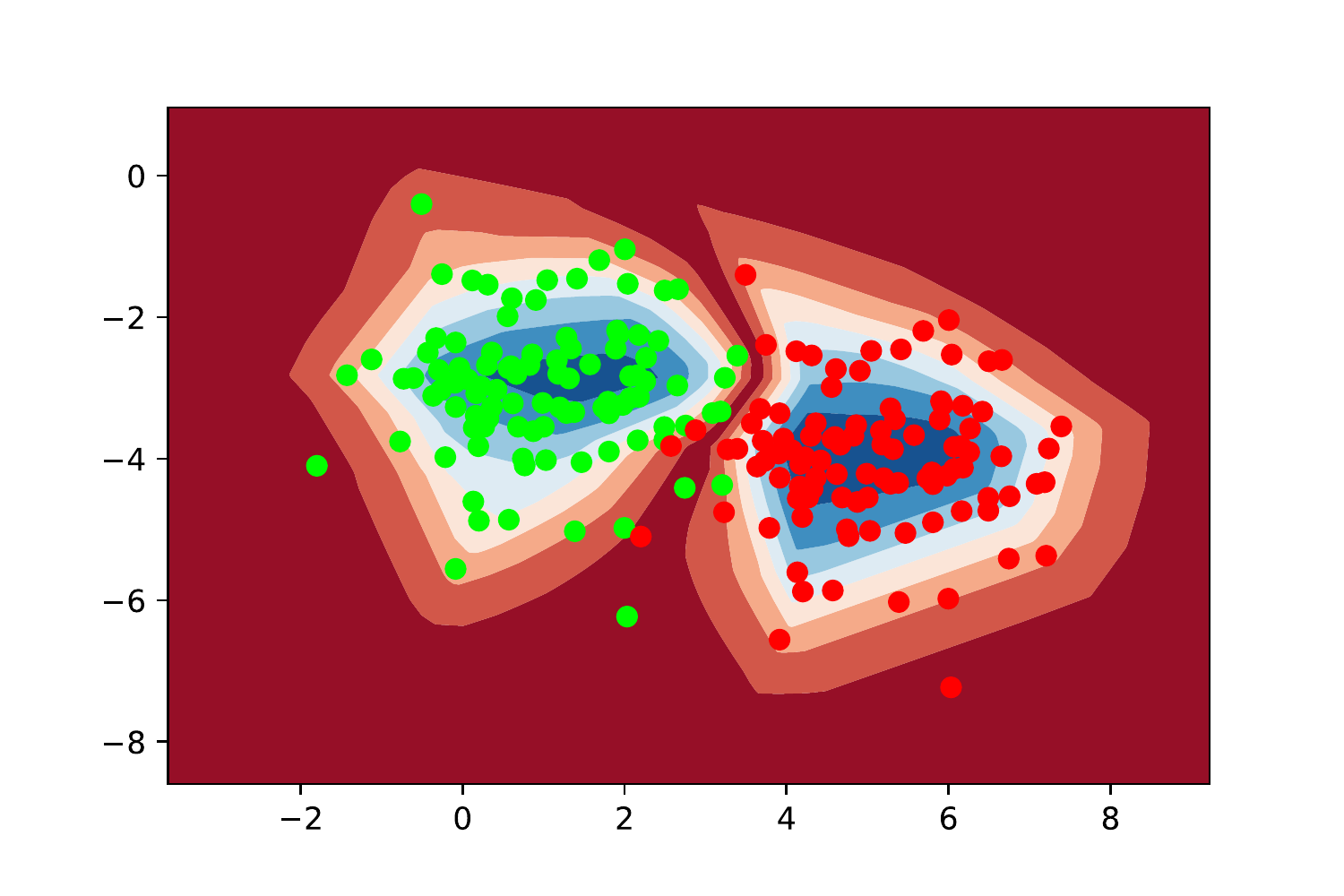} \\
      (c) Generated Points &  (d) Proposed Model
  \end{tabular}
  \caption{\label{fig:class_boundaries} Class boundaries for models on a simple 2D classification problem (green vs red dots): prediction confidence depicts on a color scale from maroon (low confidence) to blue (high confidence). The generated samples are shown as blue dots in (c) along with the original data points.}
\end{figure}

%
Aforementioned approaches may still make misleading and overconfident predictions for the samples out of the training distribution.
Figures~\ref{fig:class_boundaries}(a), (b), and (d) demonstrate predicted class boundaries for a simple 2D classification problem of green vs red dots. 
Standard deterministic neural networks (Fig.~\ref{fig:class_boundaries}(a)) do not decrease their prediction confidence when classifying samples around the class boundary, while EDL (Fig.~\ref{fig:class_boundaries}(b)) does.
%
%
However, both models have high prediction confidence when tested with out-of-distribution samples.
%
To avoid such overconfident predictions, other approaches such as~\cite{prior_nets} propose to hand-pick an auxiliary data set as the out-of-distribution samples and explicitly train the neural networks to give highly uncertain output for them.
This is often infeasible in high-dimensional real-life settings, given the very large space of possibilities.


In this paper, we propose a deterministic neural network that can effectively and efficiently estimate classification uncertainty for both in- and out-of-distribution samples.
Using a generative model, it synthesizes out-of-distribution samples close to the training samples, e.g., the blue dots in Fig.~\ref{fig:class_boundaries} (c).
Then, it trains a classifier using both training and the generated samples.
Figure~\ref{fig:class_boundaries}(d) depicts the result of our approach on our toy example: our model shows higher prediction confidence only for regions close to the training samples.

Our contribution in the paper is threefold.
(1) We consider the output of the neural network as the parameters of a Dirichlet distribution with uniform prior, instead of a categorical distribution over possible labels.
(2) These parameters are calculated by learning an implicit density estimation for each category by treating each output of the network as an output of a binary classifier, which learns to discriminate samples of the category from the samples of other categories and out-of-distribution samples.
(3) We also propose a novel generative adversarial network, which learns to distort the training samples to automatically generate the most informative out-of-distribution samples during training, and so overcoming the need to hand-pick an auxiliary data set as the out-of-distribution samples.
%
%
%

Through extensive experiments, we compare our model not only with state-of-the-art Bayesian networks and other models for uncertainty estimation, but also with recent anomaly detection models, which are specifically designed to determine out-of-distribution samples using deep neural networks.
Our experiments on MNIST and CIFAR10 data sets and adversarial examples indicate that our approach outperforms the existing approaches significantly in these tasks.

\section{Generative Evidential Neural Networks}
\label{sec:GEN}
Our work can be considered as an extension of approaches for classification, which take the output of a neural network for an input sample to estimate parameters of a Dirichlet distribution~\cite{edl,prior_nets,gast2018lightweight} for its classification.
That is, the resulting Dirichlet distribution represents the likelihood of each possible categorical distribution over the labels for the classification of the sample.

More formally, the Dirichlet distribution is a probability density function (pdf) for possible values of the probability mass function (pmf) $\mathbf{p}$.
It is characterized by $K$ parameters $\bm{\alpha}=[\alpha_1, \cdots, \alpha_K]$ and is given by
\begin{equation}
D(\mathbf{p}|\bm{\alpha}) = \left \{ \begin{array}{ll}
\frac{1}{B(\bm{\alpha})} \prod_{i=1}^K p_i^{\alpha_i-1} & \mbox{for $\mathbf{p} \in \mathcal{S}_K $},\\
0 & \mbox{otherwise}, \end{array} \right .
\end{equation}
where $\mathcal{S}_K$ is the $K$-dimensional unit simplex and $B(\bm{\alpha})$ is the $K$-dimensional multinomial beta function~\cite{kotz.00}.

In classical neural networks for classification, softmax function is used to predict class assignment probabilities.
However, it provides only a point estimate for the class probabilities of a sample and does not provide the associated uncertainty for this prediction.
On the other hand, Dirichlet distributions can be used to model a probability distribution for the class probabilities.
For instance, a Dirichlet distribution whose all parameters are one, i.e., $D(\mathbf{p}|\langle 1,\ldots,1 \rangle)$ or shortly $D(\bm{p} |\bm{1})$, represents the uniform distribution over all possible assignment of class probabilities and means total uncertainty for the classification of a sample.
As the parameter referring to specific class increases, the likelihood of probability assignments with higher values for this class also increases.
For instance, $D(\mathbf{p}|\langle 2,\ldots,1 \rangle)$ indicates that probability distributions placing more mass on the first class are slightly more likely, while their likelihood increases further for $D(\mathbf{p}|\langle 10,\ldots,1 \rangle)$, which indicates that $8$ more pieces of evidence is observed for the assignment of the sample to the first class~\cite{SLbook}.

The parameters of a Dirichlet distribution are associated with pseudocounts representing the number of observations or evidence in each class.
Hence, the predicted Dirichlet distribution for a sample may refer to the amount of evidence observed on the training set for the assignment of the sample to classes.
If there is no evidence for the assignment, we consider a 
uniform prior, i.e., $D(\mathbf{p}|\langle 1,\ldots,1 \rangle)$ and any evidence $e_i$ for class $i$ should add up to the relevant parameter of this prior (i.e., $\alpha_i = 1 + e_i$) to generate the predicted Dirichlet distribution for the sample.
The mean and the variance of a Dirichlet distribution for the class probability $p_k$ are computed as
\begin{equation}
\label{e:pignistic}
\hat{p}_k = \frac{\alpha_k}{S}~~ \mbox{and} ~~ Var(p_k) = \dfrac{\alpha_{k} (S-\alpha_{k})}{ S^2 (S + 1) },
\end{equation}
where $S=\sum_{i=1}^K \alpha_k$.
The Dirichlet distribution after incorporation of a number of evidence, $\hat{\mathbf{p}}$ represents the minimum mean square error (mmse) estimate of the ground truth appearance probabilities given these observations.

In this paper, we consider  Dirichlet distributions with uniform prior, which means that $S \geq K$. Then, the total evidence used to update the uniform Dirichlet to  the predicted Dirichlet distribution becomes $S-K$.
After predicting the parameters of the Dirichlet distribution for each sample, previous approaches have used its mean, i.e., $\hat{\mathbf{p}}$, as the class assignment probabilities for decision making, while using $K/S \in [0-1]$ as the associated uncertainty of this assignment~\cite{edl}.
As the total evidence increases, the variance of the Dirichlet distribution decreases, and so does the uncertainty of prediction.

In this work, we also use the mean of Dirichlet distributions as the predictive categorical distribution for classification.
However, to be consistent with literature in general and for benchmark comparisons, we adopt the entropy of class probabilities as a proxy for classification uncertainty~\cite{gal2016icml,Louizos17}.
%
%

\subsection{Learning to Quantify Classification uncertainty}
\label{sec:uncertainty}
Recent approaches also used Dirichlet distribution to quantify classification uncertainty in deep neural nets.
However, they failed to link the calculated Dirichlet parameters to the observations or the evidence derived from the distribution of the training set.
That is why these models could still be able to derive large amount of evidence and become overconfident in their predictions for out-of-distribution samples.

%
%


We use ideas from implicit density models~\cite{mohamed2016learning} and noise-contrastive estimation (NCE)~\cite{gutmann2012noise} to derive Dirichlet parameters for samples.
Let us consider a classification problem with $K$ classes and assume $P_{in}$, $P_k$, and $P_{out}$ represent respectively the data distributions of the training set, class $k$, and out-of-distribution samples, i.e., the samples that do not belong to any of $K$ classes.
A convenient way to describe density of samples from a class $k$ is to describe it relative to the density of some other reference data.
By using the same reference data for all classes in the training set, we desire to get comparable quantities for their density estimations.
In NCE, noisy training data~\cite{hafner2018reliable} is usually used as reference, here, we generalize this as the out-of-distribution samples, which may also include the noisy data.

Using the dummy labels $y$, we can reformulate the ratio of the densities $P_k(\bm{x})$ and $P_{out}(\bm{x})$ for a sample $\bm{x}$ as follows:
\begin{equation}
\label{eq:log_ratio}
\frac{P_k(\bm{x})}{P_{out}(\bm{x})} = \frac{p(\bm{x}|y = k)}{p(\bm{x}|y=out)} = \frac{p(y=k|\bm{x})}{p(y=out|\bm{x})} \left(\frac{1-\pi_k}{\pi_k} \right)
\end{equation}
where $\pi_k$ is the marginal probability $p(y=k)$ and $(1-\pi_k)/\pi_k$ can be approximated as the ratio of sample size, i.e., $n_k/n_{out}$, which is taken as one in this work for simplicity without loss of generality.

As shown in Eq.~\ref{eq:log_ratio}, one can approximate the log density ratio $\log\big({P_k(\bm{x})}/{P_{out}(\bm{x})}\big)$ as the logit output of a binary classifier~\cite{mohamed2016learning}, which is trained to discriminate between the samples from $P_k$ and $P_{out}$.
Let a neural network classifier $\bm{f}(\bm{x} | \theta)$ parameterized by weights $\theta$ have $K$ outputs for a given sample $\bm{x}$, where each output $f_k(\bm{x} | \theta)$ corresponds to a logit for one of $K$ classes and approximates $\log\big({P_k(\bm{x})}/{P_{out}(\bm{x})}\big)$.
To train such a network, we use the Bernoulli (logarithmic) loss as given by 
\begin{equation}
\label{eq:L1}
\begin{multlined}
\mathcal{L}_1(\theta) = -\sum_{k=1}^K  \Big[ \E_{P_k(\bm{x})} [log( \sigma( f_k(\bm{x}|\theta)))] +\\ \E_{P_{out}(\bm{x})} [log(1- \sigma( f_k(\bm{x}|\theta)  )  )] \Big].
\end{multlined}
\end{equation}

The expectations in Eq.~\ref{eq:L1} are computed by Monte Carlo integration using the equal number of samples from $P_k$ and $P_{out}$.
In this work, we use samples from $P_{out}$, which are generated by perturbing training set using a novel generative adversarial network.
The generated samples are separable from the training samples in high dimensional input space, so they are out of distribution, while still having many similarities to the training samples in a lower dimensional representation space, which is learned to reconstruct training samples.
%

As a result, $exp(f_k(\bm{x} | \theta))$ approximates the relative density $P_k(\bm{x}) / P_{out}(\bm{x})$ of class $k$, as it is trained using the samples from class $k$ and the samples close to, but easily differentiable from, the samples belonging to all $K$ classes in the training set.
For each sample $\bm{x}$ in the training set, we take $\bm{e} = exp(\bm{f}(\bm{x} | \theta))$ as the pseudocounts (i.e., evidence) vector, where each element $e_{k} = exp(f_k(\bm{x} | \theta))$ is the pseudocount for $\bm{x}$ being assigned to class $k$.
Then, the parameters of the  Dirichlet distribution given the uniform Dirichlet prior is calculated as $\bm{\alpha} = \bm{e} + \bm{1} $.
If the sample $x$ is more similar to the samples from $P_{out}$, then almost zero evidence is generated by the neural network and the predicted Dirichlet distribution becomes very close to the uniform Dirichlet distribution.
This should be the case for samples from $P_{out}$ and for the outliers in the training set.
On the other hand, if the sample is labelled as $k$ in the training set and it is not an outlier, we expect $e_{k} > e_{j} \geq 0$ for any $j \neq k$.

\subsection{Uncertainty for Misclassified Samples}
The computed $\bm{\alpha}$ parameters defines a Dirichlet distribution $D(\bm{p} | \bm{\alpha})$, from which one can sample categorical distributions over possible classes of $\bm{x}$.
However, only one of these classes is correct and assignment of $\bm{x}$ to any other class is considered as a misclassification. If $k$ is the true class of $\bm{x}$, then the marginal distribution for $p_k$, i.e., the probability of correctly classifying $\bm{x}$, is a two-parameter Dirichlet distribution (also known as Beta distribution) with parameters $\langle {\alpha}_k, \sum_{j \neq k} {\alpha}_j \rangle$.
Let $\bm{p}_{-k}$ refer to the vector of probabilities $p_j$ such that $j \neq k$.
Probabilities of misclassifying $\bm{x}$ to each class other than the true one is distributed based on the conditional Dirichlet distribution $\bm{p}'_{-k} | p_k \sim  D(\bm{p}'_{-k} | \bm{\alpha}_{-k})$, where $\bm{p}'_{-k}$ is a categorical distribution over misclassified classes and  created by normalizing $\bm{p}_{-k}$ with $(1-p_k)$, which is also equivalent to $\sum_{i \neq k} p_i$. %
%
%
Since we desire a classifier to be totally uncertain in its misclassifications (except near decision boundaries), we minimize Kullback–Leibler ($\mathbb{KL}$) divergence between $D(\bm{p}'_{-k} | \bm{\alpha}_{-i})$ and the uniform Dirichlet distribution $D(\bm{p}_{-k} |\bm{1})$ using the following regularizer:
\begin{equation}
\label{eq:reg_loss}
\mathcal{L}_2(\theta | \bm{x} ) = \beta \mathbb{KL} [D(\bm{p}_{-k}| \bm{\alpha}_{-k}) \mid\mid D(\bm{p}_{-k} |\bm{1}) ],
\end{equation}
where $\beta$ is the weight of the $\mathbb{KL}$ term. It can be set to $(1-\hat{p}_{k})$, which is the expectation for the probability of misclassification (i.e., $1-p_k$) and its usage as a weight of the $\mathbb{KL}$ term enables the learned loss attenuation; that is, it places a higher weight on epistemic uncertainty enforcement 
as the aleatoric uncertainty for misclassification decreases. 
%
%
%
%
Then, the generative evidential neural network learns the parameters $\theta$ by minimizing the overall loss defined as
\begin{equation}
\mathcal{L}(\theta) = \mathcal{L}_1(\theta) + \E_{P_{in}(\bm{x})} [\mathcal{L}_2(\theta | \bm{x})].
\end{equation}
%

\subsection{Generating out-of-distribution samples}
\label{sec:gan}
Previous approaches generate out-of-distribution samples by perturbing training samples. However, they usually require manual determination of how much perturbation should be made to these samples.
Too little perturbation may not set the resulting samples apart 
from the actual samples, while too much perturbation may cast them far away from the training data and deteriorate their usefulness.

Variational Autoencoders (VAE) are probabilistic generative models that create low dimensional latent representations for high dimensional data by maximizing 
\begin{equation}
\label{eq:vae}
\small
\begin{multlined}
\max_{q_\theta,p_\phi
} \sum_{i=1}^N \mathbb{E}_{q_\theta(\bm{z} \mid \bm{x}_i)} \big [ \log p_\phi(\bm{x}_i \mid \bm{z})  - \mathbb{KL}(q_\theta(\bm{z}\mid \bm{x}_i) \mid\mid p(\bm{z})) \big],
\end{multlined}
\end{equation}
where $q_\theta(\bm{z}\mid \bm{x}_i)$ is the latent space distribution for each sample $\bm{x_i}$ and $p_\phi(\bm{x}_i \mid \bm{z})$ is the decoder likelihood distribution that is maximized for each sample $\bm{x_i}$.  The  $\mathbb{KL}$ term enforces  $q_\theta(\bm{z}\mid \bm{x}_i)$ to be close to a prior distribution $p(\bm{z})$ and have a denser latent space. 
%
%

%
Proximity of the encoded samples in the latent space of a VAE is commonly used as an indication of their semantic similarity and exploited for few-shot classification and anomaly detection tasks.
In this work, we also use the latent space of a VAE as a proxy for semantic similarity between samples in input space.
Hence, we exploit it to generate out-of-distribution samples, which are similar to, but at the same time clearly separable from, the training examples in the input space.
%
%

\begin{figure}[t!] 
\includegraphics[width=6cm]{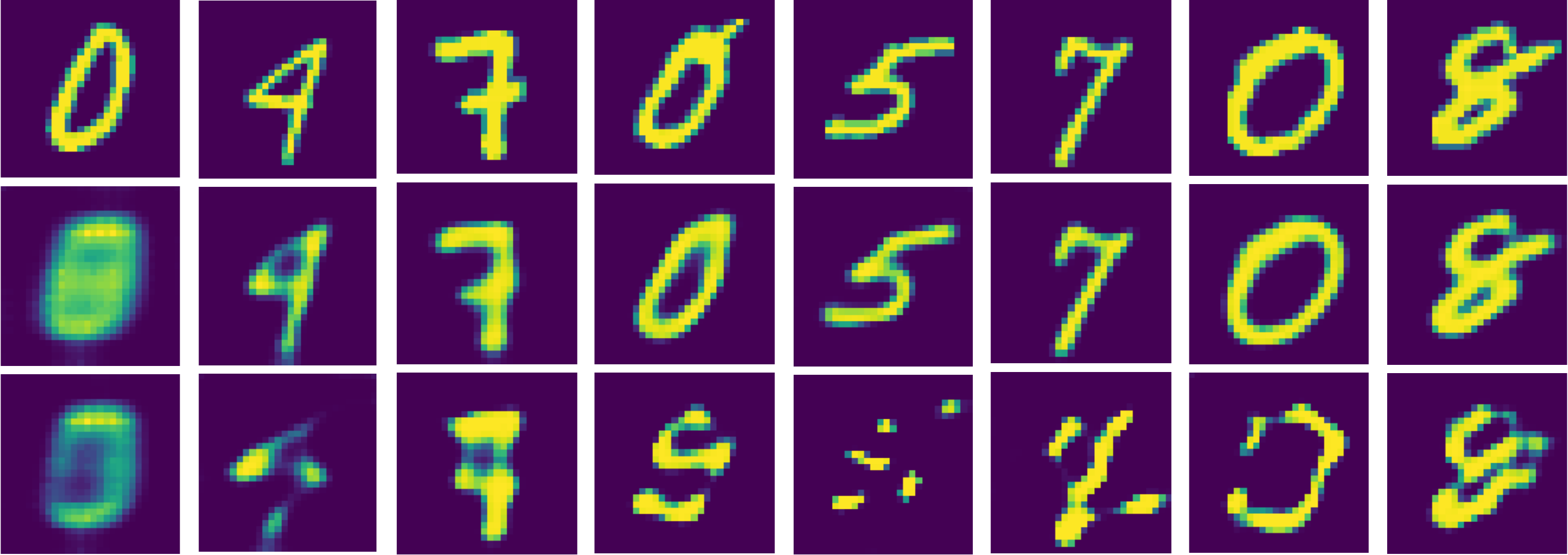}
\caption{\label{fig:adversarial_MNIST} Original training samples (top), samples reconstructed by the VAE (middle), and the samples generated by the proposed method (bottom) over a number of epochs.}
\end{figure}

For each $\bm{x}_i$ in training set, we sample a latent point $\bm{z}$ from $q_\theta(\bm{z}\mid \bm{x}_i)$ and perturb it by $\bm{\epsilon} \sim q_{\gamma}(\bm{\epsilon}|\bm{z})$, which is implemented as a multivariate Gaussian distribution $\mathcal{N}(\bm{0},G(\bm{z}))$, where $G(\cdot)$ is a fully connected neural network with non-negative output that is trained via
\begin{equation}
\small
\label{eq:G}
\max_{G} ~~ \mathbb{E}_{{\substack{q_\theta(\bm{z} \mid \bm{x}_i), \\ q_{\gamma}(\bm{\epsilon}|\bm{z}), \\ p_\phi(\bm{\bar{x}}_i \mid \bm{z} + \bm{\epsilon})}}} \big[ \underbrace{ \log D'(\bm{z}+\bm{\epsilon})}_{(a)} + \underbrace{\log (1 - D(\bm{\bar{x}}_i)}_{(b)})  \big], \\
\end{equation}
where  $\bm{\bar{x}}_i \sim p_\phi(\bm{\bar{x}}_i \mid \bm{z} + \bm{\epsilon})$ is the decoded out-of-distribution sample from the perturbed sample $\bm{z} + \bm{\epsilon}$. The discriminators $D$ and $D'$ 
%
%
%
%
are binary classifiers with \textit{sigmoid} output 
that try to distinguish real samples from the generated ones. That is, given an input, a discriminator gives as an output the probability that the sample is from the training set distribution.
In Eq.~\ref{eq:G}, (a) forces the generated points to be similar to the real latent points through making them indistinguishable by $D'$ in the latent space of the VAE and (b) encourages the generated samples to be distinguishable by $D$ in the input space.
%
The discriminators are optimized via
\begin{equation}
\small
\label{eq:D2}
\max_{D'} ~~ \log D'(\bm{z}) +
\underbrace{ \log (1 - D'(\bm{z}+\bm{\epsilon})),}_{(c)}
\end{equation}
\begin{equation}
\small
\label{eq:D1}
\max_{D} ~~ \log D(\bm{x}_i) + \log (1 - D(\bm{\bar{x}}_i)).
\end{equation}
%
Note that (c) of Eq.~\ref{eq:D2} 
is also included in 
the objective of the VAE (Eq.~\ref{eq:G}) to adapt the latent space during the training of the generator.
We trained the VAE, generator, and discriminators by iterating between maximizing Eq.~\ref{eq:vae} through Eq.~\ref{eq:D1} until convergence,
as in the regular training of generator and discriminator in GANs.
We demonstrate this approach in Fig~\ref{fig:class_boundaries} (c) and  Fig.~\ref{fig:adversarial_MNIST}, where a number of real and generated MNIST images are shown.

\begin{figure*}[t]
\CenterFloatBoxes
\begin{floatrow}
\capbtabbox{
  \resizebox{5.0cm}{!}{
  \begin{tabular}{|c|l|c|c|c|c|}
    \hline
     & \multicolumn{1}{c|}{\textbf{Layer}} & \multicolumn{1}{c|}{\textbf{Filters/Neurons}} & \multicolumn{1}{c|}{\textbf{Patch Size}} & \multicolumn{1}{c|}{\textbf{Stride}} & \multicolumn{1}{c|}{\textbf{Activation}} \\ \hline
     \multirow{3}{*}{\STAB{\rotatebox[origin=c]{90}{Classifier~~~~~~}}}
        & Conv$_1$ & 20 & 5 $\times$ 5 & 1 & relu\\
        & Max Pool & - & 2 $\times$ 2 & 2 & -\\
        & Conv$_2$ & 50 & 5 $\times$ 5 & 1 & relu\\
        & Max Pool & - & 2 $\times$ 2 & 2 & -\\
        & FC$_1$ & 500 & - & - & relu\\
        & FC$_2$ & $K=10$ & - & - & - \\\hline
     \multirow{3}{*}{{\rotatebox[origin=c]{90}{~~~$D$}}}
        & Conv$_1 -$ FC$_1$ & repeat & repeat & repeat & repeat \\
        & FC$_3$ & 1 & - & - & sigmoid \\\hline
     \multirow{3}{*}{{\rotatebox[origin=c]{90}{~~$G$~~~~~}}}
        & FC$_4$ & 32 & - & - & relu\\
        & FC$_5$ & 32 & - & - & relu\\
        & FC$_6$ & 32 & - & - & relu\\
        & FC$_7$ & $code\_sz=50$ & - & - & softplus \\\hline
     \multirow{3}{*}{{\rotatebox[origin=c]{90}{~~~~$D'$}}}
        & FC$_4 -$ FC$_6$ & repeat & repeat & repeat & repeat \\
        & FC$_8$ & 1 & - & - & sigmoid \\\hline
  \end{tabular}}
}{
  \caption{Network architectures. \label{tab:architecture}}
}
\capbtabbox{
 \resizebox{3.5cm}{!}{
  \begin{tabular}{|c|c|c|} \hline
  \textbf{Model} & \textbf{MNIST} & \textbf{CIFAR 5}\\ \hline
  {L2} & 99.4 & 76\\
  {Dropout} & 99.5 & 84 \\
  {Deep Ensemble} & 99.3 & 79 \\
  {FFG} & 99.1 & 78\\
  {FFLU} & 99.1  & 77\\
  {MNFG} & 99.3 & 84\\
  {BBH} & 99.1 & 80\\
  {EDL} & 99.3 & 83\\
  {GEN} & 99.3 & 83\\
  \hline
  \end{tabular}}
}{
  \caption{Test accuracies (\%) for MNIST and CIFAR5. \label{tab:accuracies}}  %
}
\end{floatrow}
\end{figure*}

\section{Evaluation}
\label{sec:eval}
To be able to compare our approach with the recent work, we adopted the same strategy used for evaluation in~\cite{Louizos17,edl,pawlowski2017implicit}.
That is, we use LeNet-5~\cite{lecun1998gradient} with ReLu non-linearities and max pooling as the neural network architecture and evaluated our approach with MNIST and CIFAR10 datasets, to be able to make a fair comparison with the most related recent work.
We implemented our approach
\footnote{https://muratsensoy.github.io/gen.html}
using Python and Tensorflow.

In this section, we compared our approach with the following approaches:
\begin{inparaenum}[\noindent \itshape \upshape(a\upshape)]
\item  \textbf{L2} corresponds to the standard neural nets with softmax probabilities and L2 regularization,
\item \textbf{Dropout} refers to the Bayesian model used in~\cite{gal2015bayesian},
\item \textbf{Deep Ensemble} refers to the model proposed in~\cite{lakshminarayanan2017simple},
\item \textbf{FFG} refers to the fully factorized Bayesian model with Gaussian posteriors from~\cite{blundell2015icml}, which is widely known as \textit{Bayes by Backprop} (\textbf{BBB}),
\item \textbf{FFLU} refers to the Bayesian model used in~\cite{kingma2015variational} with the additive parametrization~\cite{molchanov2017variational},
\item \textbf{MNFG}\footnote{https://github.com/AMLab-Amsterdam/MNF\_VBNN} refers to the variational approximation based model in~\cite{Louizos17},
\item \textbf{EDL} refers to the model in~\cite{edl},
\item \textbf{BBH}\footnote{https://github.com/pawni/BayesByHypernet} refers to the Bayesian model based on implicit weight uncertainty ~\cite{pawlowski2017implicit}, and \item \textbf{GEN} refers to the proposed approach.
\end{inparaenum}

\subsection{Predictive Uncertainty Estimation}
We used the network architectures in Table~\ref{tab:architecture} to train our model for the MNIST dataset. For CIFAR10, we used the same architectures; however, the classifier uses $192$ filters for Conv$_1$ and Conv$_2$, also has $1000$ neurons in FC$_1$ as described in~\cite{Louizos17}.
We used L2 regularization with coefficient $0.005$ in the fully-connected layers.
Other approaches are also trained using the same classifier architecture with the priors and posteriors described in~\cite{Louizos17} and ~\cite{pawlowski2017implicit}.
The classification accuracy of each model on the MNIST test set can be seen at Table~\ref{tab:accuracies}. While we do not explicitly aim for high classification accuracy, our results indicate that our approach is doing better than most of the other approaches.

\begin{figure*}[t!]
$\begin{array}{cc}
\includegraphics[width=5.cm]{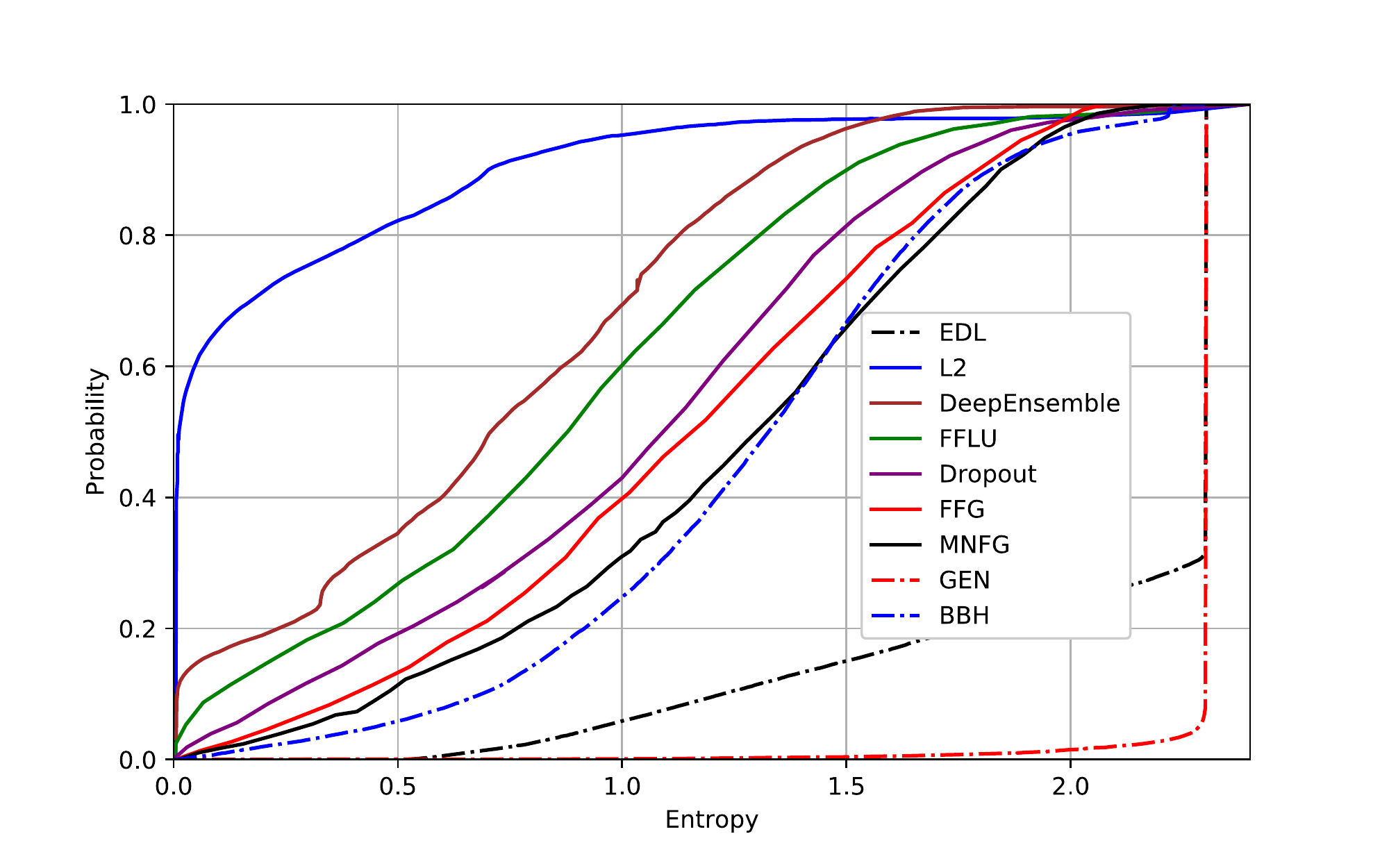} & \includegraphics[width=5.cm]{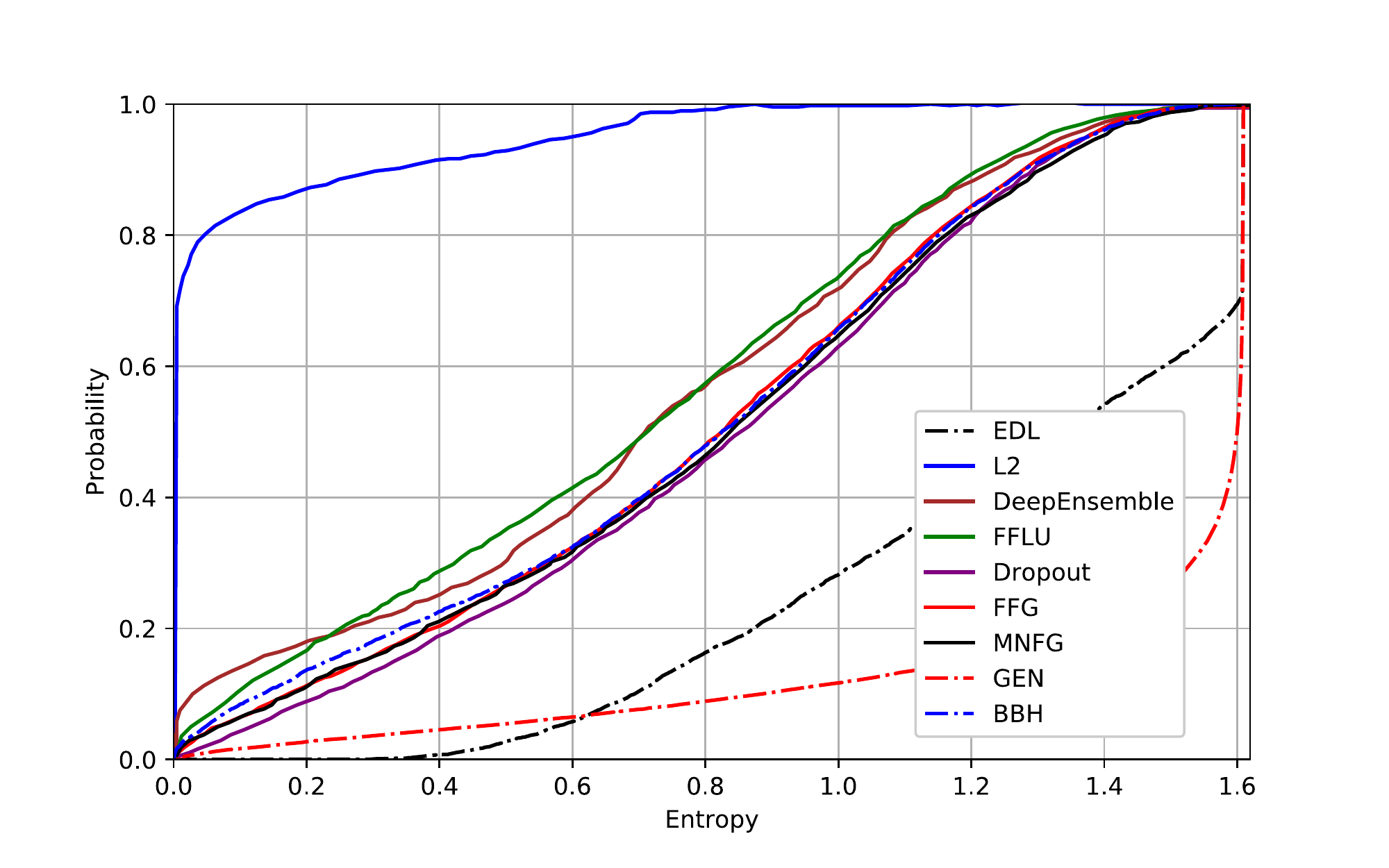}
\end{array}$
\caption{\label{fig:entropy_mnist} Empirical CDF for the entropy of the predictive distributions on the notMNIST dataset (left) and samples from the last five categories of CIFAR10 dataset (right).}
\end{figure*}

We train models for MNIST using the images from $10$ digit categories from the training set as usual. However, we then tested these models on notMNIST dataset,\footnote{https://www.kaggle.com/lubaroli/notmnist} which contains $10$ letters $A$-$J$ instead of digits.
For CIFAR10, we trained models using the training data from the first five categories (referred to as CIFAR5) and tested these models using the images from the last five categories.
For both MNIST and CIFAR10, the predicted label for any test sample is guaranteed to be wrong, since test samples are coming from a different distribution than the one for 
the training set.
Hence, an ideal classifier should 
report totally uncertain predictions instead of associating a higher likelihood for a specific label for an out-of-distribution test sample.

To be consistent with recent works, we use the entropy of the predicted categorical distribution over labels as a proxy to quantify prediction uncertainty. That is, a prediction gets more uncertain as its entropy approaches the maximum entropy, i.e., the entropy of the uniform categorical distribution.
As in~\cite{Louizos17}, we used the empirical CDF of entropy distribution for predictions to quantify how uncertain they are.
That is, as the predictions get more uncertain, the area under their entropy CDF curve gets smaller.

\begin{figure*}[t!]
\includegraphics[width=9cm, height=4.3cm]{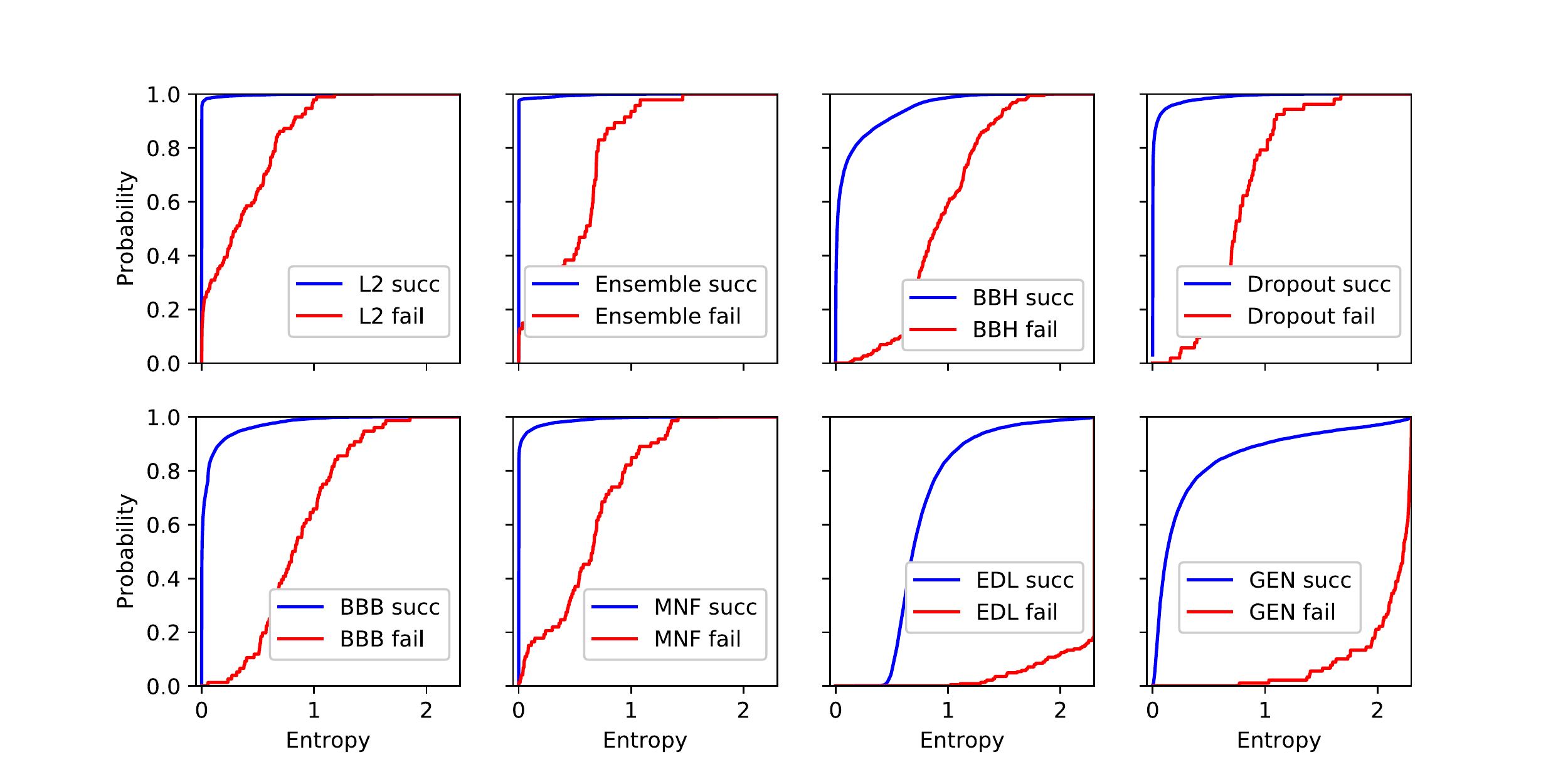}
\caption{\label{fig:MNIST_succ_fail} Entropy CDF curves of different models for their successful and failed predictions on the MNIST test set.}
\end{figure*}

Figure~\ref{fig:entropy_mnist} shows our results for MNIST and CIFAR10 datasets. Standard neural networks (referred to as L2) is very confident in its predictions as indicated by its entropy CDF curves in the figure.
On the other hand, Bayesian neural network models appear to be more uncertain about their predictions with respect to the standard neural networks.
The performances of these models in terms of predictive uncertainty vary for MNIST while they perform almost the same for CIFAR10.
EDL and GEN perform much better than Bayesian approaches in both MNIST and CIFAR10.
Especially, GEN associate very high uncertainty with its predictions for out-of-distribution samples.

After conducting these benchmark analysis by following the very same procedure proposed in ~\cite{Louizos17}, we also tested these models with in-distribution samples and analyzed the certainty they assign to correct and incorrect predictions. 
%
%
Figure~\ref{fig:MNIST_succ_fail} 
shows entropy CDF curves for successful and failed predictions in MNIST test set for different models.
The figure indicates that standard networks and Bayesian neural networks are overconfident (i.e., have low entropy) for their failed predictions; that is, they have large area under the entropy CDF curve for their failed predictions (i.e., misclassifications).
However, both EDL and GEN have significantly higher predictive uncertainty for their failed predictions.
Furthermore, GEN gives a better disparity between the successful and failed predictions in terms of uncertainty.
We also conducted the same analysis for the CIFAR10 dataset and 
obtained similar results.

\subsection{Robustness to Adversarial Examples}
Robustness to adversarial examples is an important challenge for machine learning models.
While it is very hard to provide correct predictions for carefully crafted adversarial examples, a model should associate very high uncertainty with its prediction when tested on them. 
Hence, in this section, we test different models using a well-known white-box attack strategy, the Fast Gradient Sign Method (FGSM), proposed by~\cite{goodfellow2014explaining} and analyze how uncertain these models are when they fail to correctly classify the generated adversarial examples.

\begin{figure*}[t!]
$\begin{array}{cc}
\includegraphics[width=5cm]{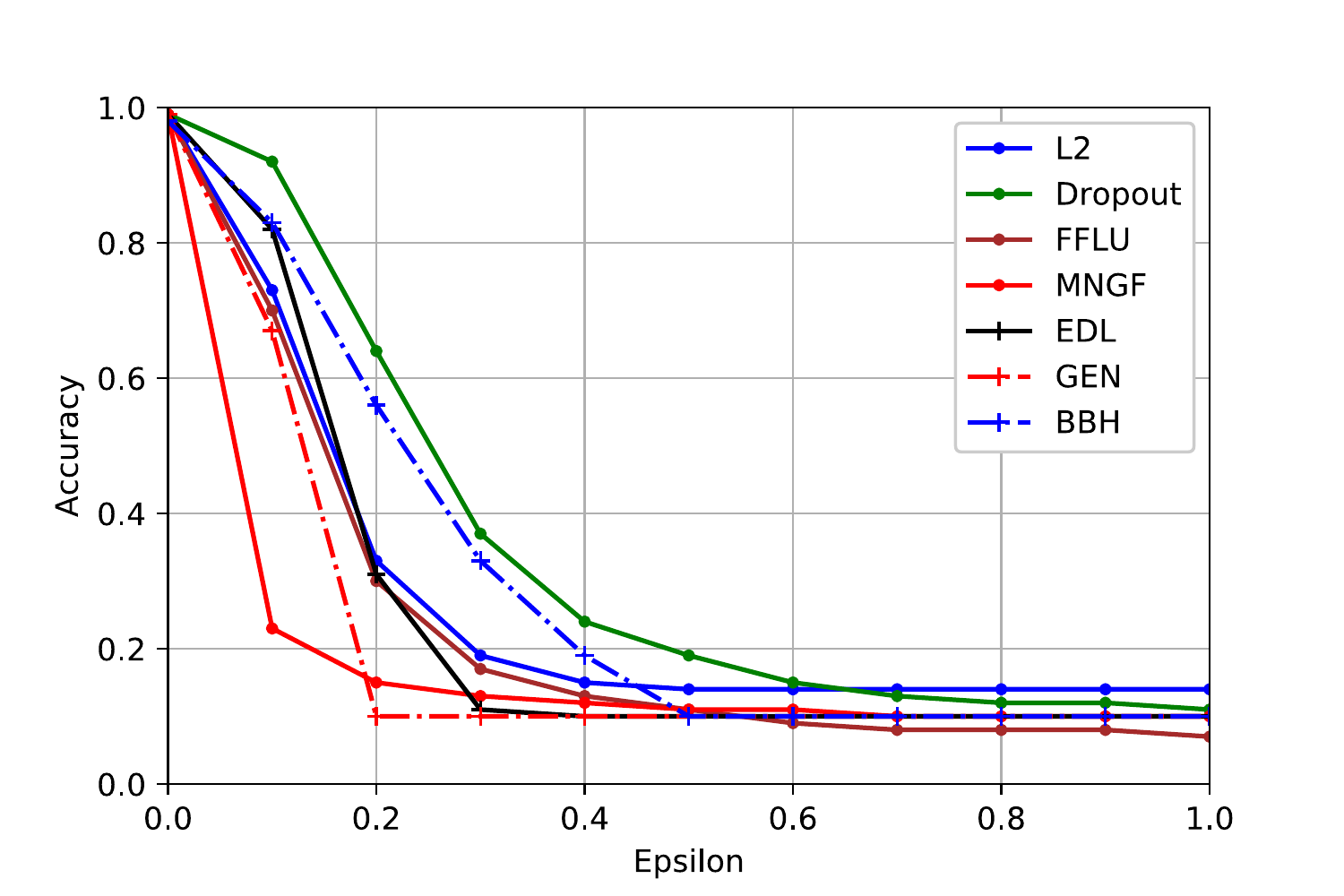} & \includegraphics[width=5cm]{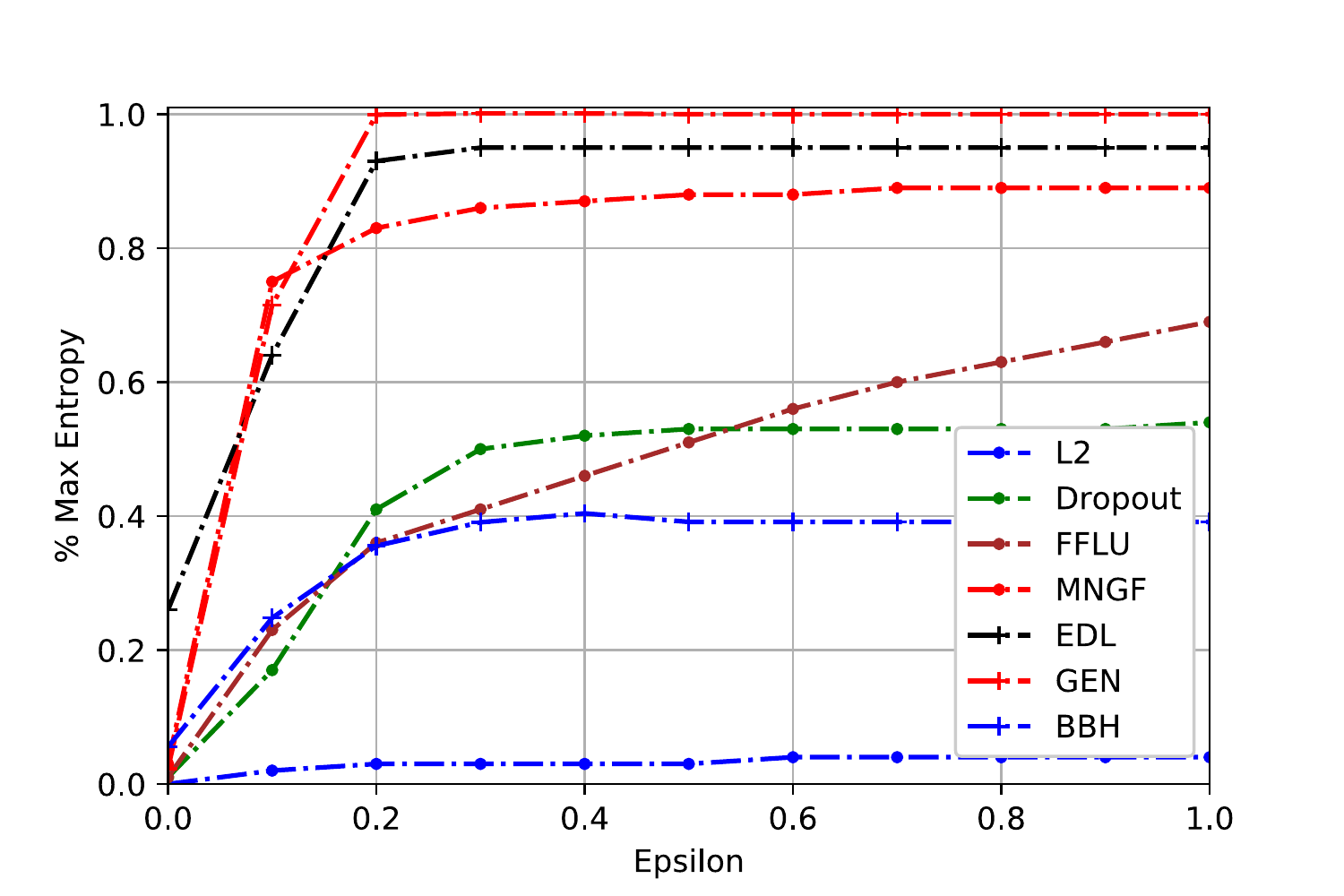}
\end{array}$
\caption{\label{fig:adversarial_MNIST_results} Accuracy and entropy as a function of the adversarial perturbation $\epsilon$ on the MNIST dataset.}
\end{figure*}

\begin{figure*}[t!]
$\begin{array}{cc}
\includegraphics[width=5cm]{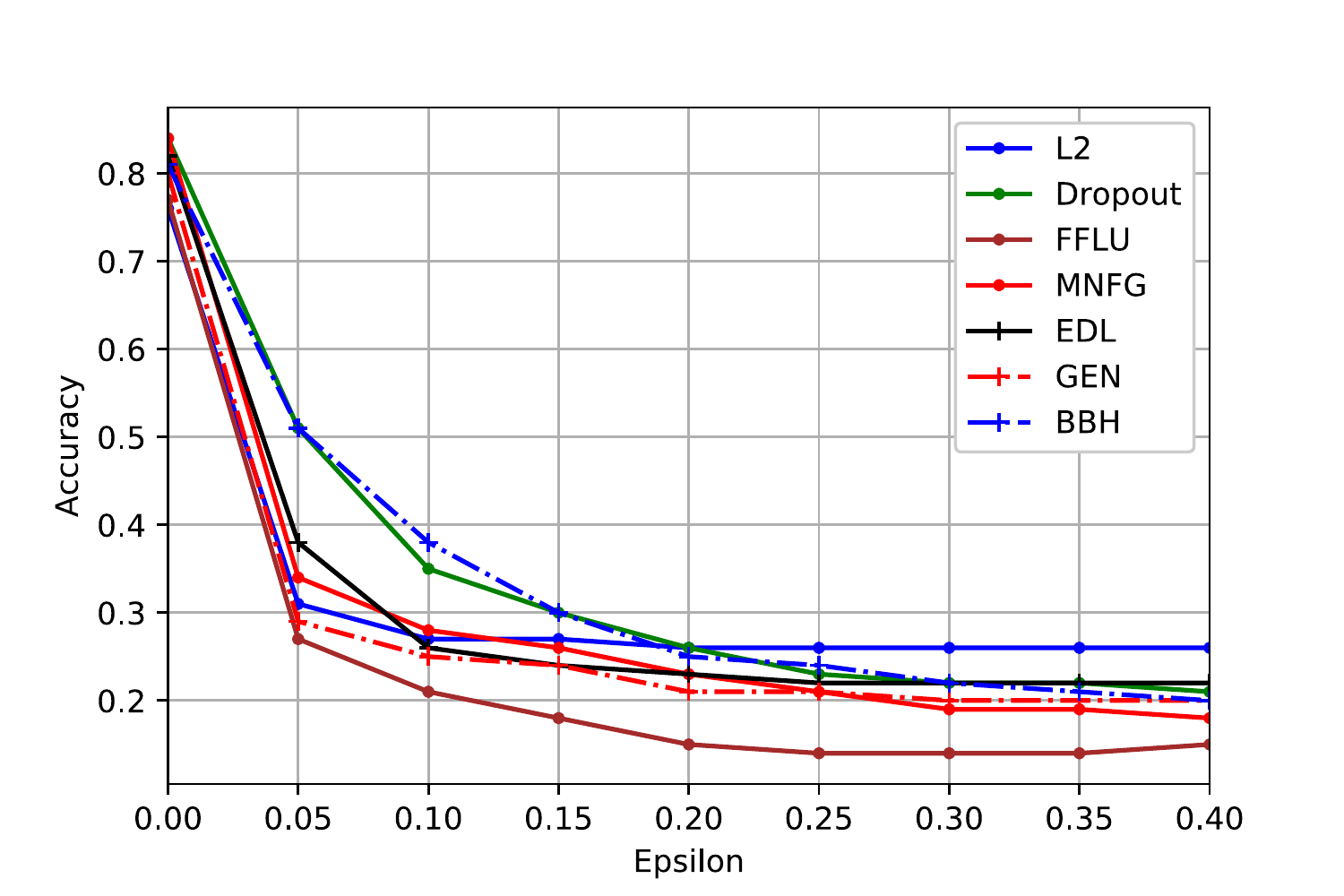} & \includegraphics[width=5cm]{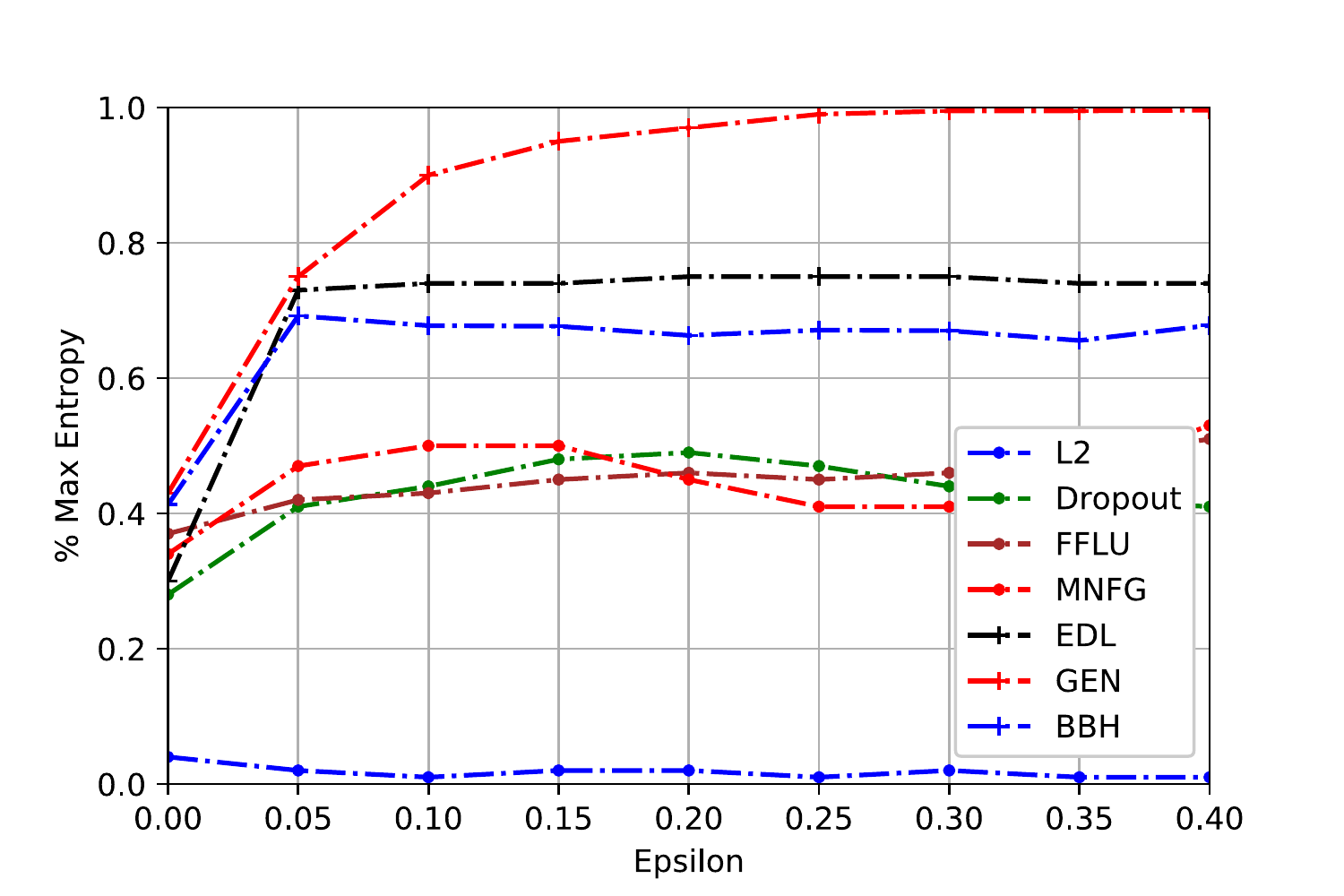}
\end{array}$
\caption{\label{fig:adversarial_CIFAR} Accuracy and entropy as a function of the adversarial perturbation $\epsilon$ on the CIFAR10 dataset.}
\end{figure*}

White-box attacks have access to model parameters and exploit gradients of the loss with respect to an input to perturb the input to create an adversarial example.
The amount of perturbation is defined by the $\epsilon \in [0,1]$ parameter.
Figure~\ref{fig:adversarial_MNIST_results} shows our results in terms of both accuracy and uncertainty for the MNIST test set for different $\epsilon$ values.
The figure indicates that GEN demonstrates the ideal behavior; it associates the highest uncertainty (maximum entropy) with its predictions as it starts to fail making the right predictions for high values of $\epsilon$.
We observe the same behavior for CIFAR10 dataset as shown in Figure~\ref{fig:adversarial_CIFAR}.

\subsection{Comparisons with Anomaly Detection Methods}
Our work is also related to anomaly detection approaches, which are specifically designed to detect out-of-distribution samples.
Hence, in this section, we compare our approach with the existing and the most recent anomaly detection methods on MNIST and CIFAR10 datasets.
We compared our approach with the following models:
\begin{inparaenum}[\noindent \itshape \upshape(a\upshape)]
\item \textbf{Calibrated} refers to the calibration-based model for out-of-distribution detection in~\cite{lee2018training},
\item \textbf{GEOTRANS} is the anomaly detection model in~\cite{golan2018deep},
\item \textbf{SVM} refers to the one-class SVM applied to the latent space of convolutional autoencoder as described in~\cite{golan2018deep},
\item \textbf{ADGAN} is the anomaly detection method based on generative adversarial networks in~\cite{deecke2018anomaly},
\item \textbf{DAGMM} is the deep autoencoding Gaussian mixture model in ~\cite{zong2018deep},
\item \textbf{DSEBM} is the deep structured energy-based model in~\cite{Zhai2016}.
\end{inparaenum}
We use publicly available implementations of the {Calibrated}\footnote{https://github.com/alinlab/Confident\_classifier} and {GEOTRANS}\footnote{https://github.com/izikgo/AnomalyDetectionTransformations} by their authors, which also contains implementations of other models above.
Unlike our approach, these models predicts a \textit{score} for out-of-distribution classification.
To evaluate these approaches, the area under the ROC curve (AUC) is used as a measure of how well the produced scores can distinguish between in- and out-of-distribution samples.

\begin{figure*}[t!]
$\begin{array}{cc}
\includegraphics[width=5cm]{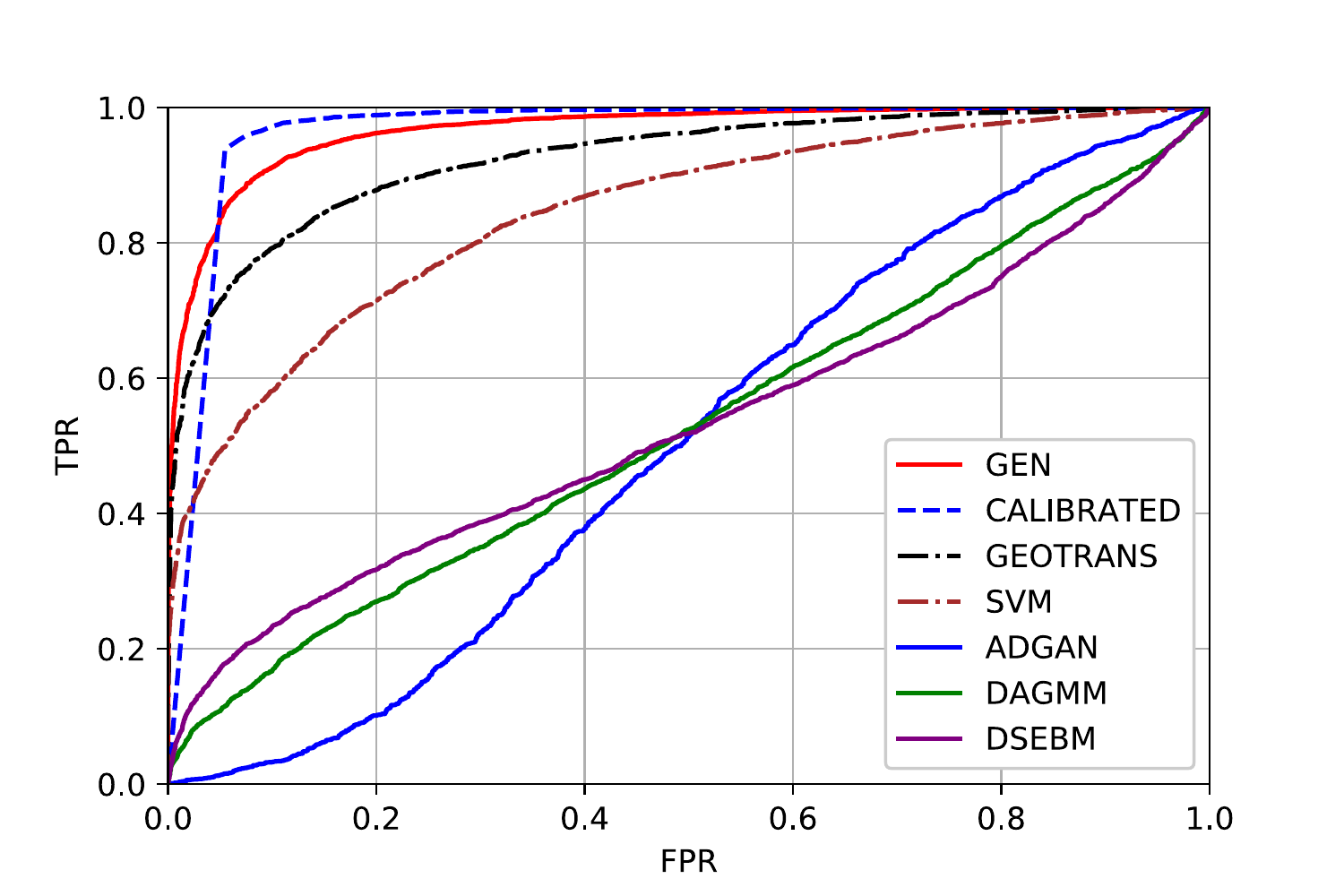} &  \includegraphics[width=5cm]{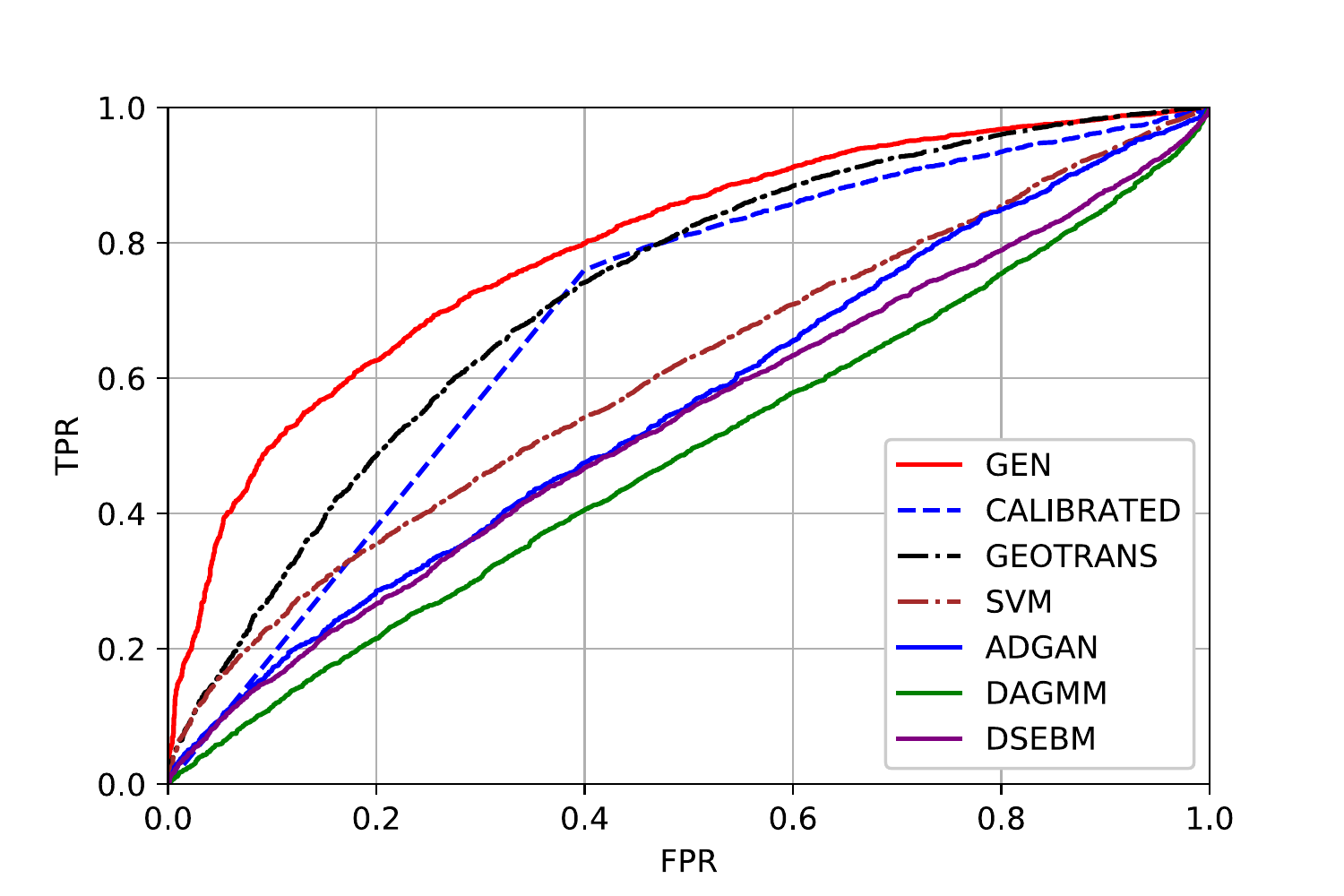}
\end{array}$
\caption{\label{fig:anomaly} Anomaly detection results for MNIST (left) and CIFAR10 (right).}
\end{figure*}

Similarly, as before, 
we train models using samples only from the first five categories of the MNIST and CIFAR10 datasets. Then, we evaluate these models on the test samples half of which comes from the first five and other half comes from the last five categories.
We use the entropy of the predictive probabilities from our model as a \textit{score} to differentiate between in- and out-of-distribution samples.
Let us note that, as shown before, our approach provides highly uncertain predictions not only for out-of-distribution samples, but also for the misclassified in-distribution samples. Hence, the in-distribution samples laying on the class boundary of first five categories may also be classified as out-of-distribution samples based on their entropy-based score.

Figure~\ref{fig:anomaly} shows our results with anomaly detection mothods on MNIST and CIFAR10 datasets. Both Calibrated and GEOTRANS perform better than SVM, ADGAN, DAGMM, and DSEBM in our experiments.
For MNIST, GEN and Calibrated achieve the best AUC values, respectively $0.965$ and $0.966$. 
For CIFAR10, GEN achieves the best AUC ($0.775$), and performs significantly better than state-of-the-art anomaly detectors in recognizing out-of-distribution samples.
While our approach has higher entropy for its predictions for both out-of-distribution samples and misclassified in-distribution samples,\footnote{Table~\ref{tab:accuracies} indicates 17\% of CIFAR5 test samples are misclassified.} its entropy-based score is still performing at least as good as the state-of-the-art anomaly detection methods.

\section{Related Work}
\label{sec:related}
Quantification of predictive uncertainty has always been very important for machine learning models.
Gaussian Processes (GPs) \cite{rasmussen2006mit} has been very good both in making accurate predictions and estimate their predictive uncertainties. However, these kernel-based non-parametric models cannot easily deal with high-dimensional data such as images due to the curse of dimensionality.

In recent years, Bayesian deep learning has emerged as a field combining deep neural networks with Bayesian probability theory, which provides a principled way of modeling uncertainty of machine learning models by employing prior distribution on their parameters and inferring the posterior distribution for these parameters using approximations such as Variational Bayes~\cite{blundell2015icml,gal2016icml}.
Then, the posterior predictive distribution is approximated with sampling methods, which brings a significant computational overhead and leads to noise in predictive uncertainty estimates.

In these models, predictive uncertainty is modelled by taking samples from the posterior distributions of model parameters and using the sampled parameters to create a distribution of predictions for each input of the network.
However, as we show in our experiments, modeling uncertainty of network parameters may not necessarily lead to good estimates of the predictive uncertainty of neural networks~\cite{hafner2018reliable}.
This is the case especially for the misclassified in-distribution samples, where Bayesian models associate similar levels of uncertainties with their successful and failed predictions.

Recently, a number of approaches~\cite{edl,prior_nets} have been proposed to use outputs of neural networks to estimate the parameters of the Dirichlet prior of the categorical distribution for classification, instead of predicting a categorical distribution through the \textit{softmax} function.
Then, the resulting Dirichlet distribution is used to calculate the predictive uncertainty for classification.
While similar in principle, our work distinguishes from this line of work in two folds: (1) it relates the parameters (i.e., the pseudo counts) of the resulting Dirichlet distribution to the density of the training data through noise constructive estimation, (2) it automatically synthesizes out-of-distribution samples sufficiently close to the training data, instead of hand-picking an auxiliary dataset.

Previous approaches used manually-tuned noise~\cite{hafner2018reliable} or GAN in the input space~\cite{lee2018training} to create out-of-distribution samples.
The approaches based on GAN may suffer from the so-called \textit{mode collapse} problem. To avoid it in this work, we created samples by automatically perturbing each training example in the latent space separately.
Also, to avoid generating samples too similar to or different from training examples, we used a generator with joint objectives defined over outputs of two discriminators.

\section{Conclusions}
\label{sec:conclusions}
In this work, we proposed to combine ideas from implicit density models, noise constructive density estimation, and evidential deep learning in a novel way to quantify classification uncertainty in neural networks.
We also proposed to generate out-of-distribution samples by combining the strengths of VAEs and GANs. The generated examples are used for learning an implicit density model of the training data, which is then utilized to generate pseudocounts (i.e., evidence) for Dirichlet parameters.
Through extensive experiments with well-studied datasets and comprehensive comparisons with recent approaches, we show that our approach significantly enhances the state of the art in two uncertainty estimation benchmarks: i) detection of out-of-distribution samples, and ii) robustness to adversarial examples.

\section{Acknowledgments}
This research was sponsored by the U.S. Army Research Laboratory (ARL) and the U.K. Ministry of Defence under Agreement Number W911NF-16-3-0001. The views and conclusions contained in this document are those of the authors and should not be interpreted as representing the official policies, either expressed or implied, of the U.S. Army Research Laboratory, the U.S. Government, the U.K. Ministry of Defence or the U.K. Government. The U.S. and U.K. Governments are authorized to reproduce and distribute reprints for Government purposes notwithstanding any copyright notation hereon. %
Also, Dr. Sensoy thanks to ARL for its support under grant
W911NF-16-2-0173, and Newton-Katip Çelebi Fund and TUBITAK for their support under grant 116E918.

\bibliography{ref}
\bibliographystyle{aaai}

\end{document}